\documentclass{bmvc2k}

\title{Highway Driving Dataset \\for Semantic Video Segmentation}

\addauthor{Byungju Kim}{feidfoe@kaist.ac.kr}{0}
\addauthor{Junho Yim}{junho.yim@kaist.ac.kr}{0}
\addauthor{Junmo Kim*}{junmo.kim@kaist.ac.kr}{0}

\addinstitution{
 School of Electrical Engineering \\
 Korea Advanced Institute of Science\\
 and Technology (KAIST),\\
 South Korea
}

\runninghead{Kim, Yim, and Kim}{Highway Driving Dataset}

\begin{document}

\maketitle

\begin{abstract}

Scene understanding is an essential technique in semantic segmentation.
Although there exist several datasets that can be used for semantic segmentation, they are mainly focused on semantic image segmentation with large deep neural networks.
Therefore, these networks are not useful for real time applications, especially in autonomous driving systems. 
In order to solve this problem, we make two contributions to semantic segmentation task.
The first contribution is that we introduce the semantic video dataset, the Highway Driving dataset, which is a densely annotated benchmark for a semantic video segmentation task. 
The Highway Driving dataset consists of 20 video sequences having a 30Hz frame rate, and every frame is densely annotated.
Secondly, we propose a baseline algorithm that utilizes a temporal correlation.
Together with our attempt to analyze the temporal correlation, we expect the Highway Driving dataset to encourage research on semantic video segmentation.
%In addition to analyzing the temporal correlation, we expect the Highway Driving dataset to encourage research on semantic video segmentation

\end{abstract}

%-------------------------------------------------------------------------

\section{Introduction}
Recent advance in convolutional neural networks (CNNs), which started from image classification, have resulted in great improvements in the majority of computer vision tasks.
Thus, their applications have evolved to become more complex and advanced, and they thus require deeper scene understanding.
Among the numerous computer vision tasks, we tackle the problem of semantic video segmentation for a driving scenario.
Semantic segmentation is fundamentally a classification task.
What differentiates semantic segmentation from image classification is that semantic segmentation requires class prediction for entire pixels in the given image.
Therefore, the semantic segmentation task also requires a subtle understanding of \textit{local} relationships whereas image classification focuses on abstracting the given image \textit{globally}.

Recent research on semantic segmentation is focused on images rather than videos.
Various approaches have been proposed for semantic segmentation~\cite{drn,fcn,pspnet,dilation}, and they have been successfully applied for segmenting images.
In order to improve the semantic image segmentation performance, the network architectures have been made wider and deeper.
The residual network~\cite{resnet} has been commonly used as a feature extraction module, and additional modules that are specialized in semantic segmentation are supplemented subsequently.
However, it is still challenging compared to the human-level, since small region and rare objects remain troublesome.
In particular, their application to driving scenario is in demand, as autonomous driving is an application that would directly benefits from semantic segmentation~\cite{NextSegmPredICCV17,7780714,DBLP:journals/corr/NilssonS16}.
The two major requirements for autonomous driving are reliability and real-time computation.
They are complementary to each other as reliability can be interpreted as the characteristic of \textit{having no delay in decision making}.
However, they also have an adversarial relationship as real-time computation often implies small and therefore, less powerful networks in CNN-based algorithms.

In the literature, it is known that wider and deeper networks exhibit a more reliable performance~\cite{wrn,han2017deep}.
However, they are not purely beneficial from the standpoint of semantic video segmentation.
In the case of video segmentation, time consumption should be taken into consideration during the design of an algorithm because there exists a time limit for segmenting each frame.
The size of the networks would be compromised in order avoid delaying the segmentation of the following frames.
Although several studies on the time budget have been published recently~\cite{BudgetAwareDS,icnet}, the runtime constraint is still underestimated in terms of its importance in real-world applications.

\begin{figure}[t]
\centering
\includegraphics[width=0.9\textwidth]{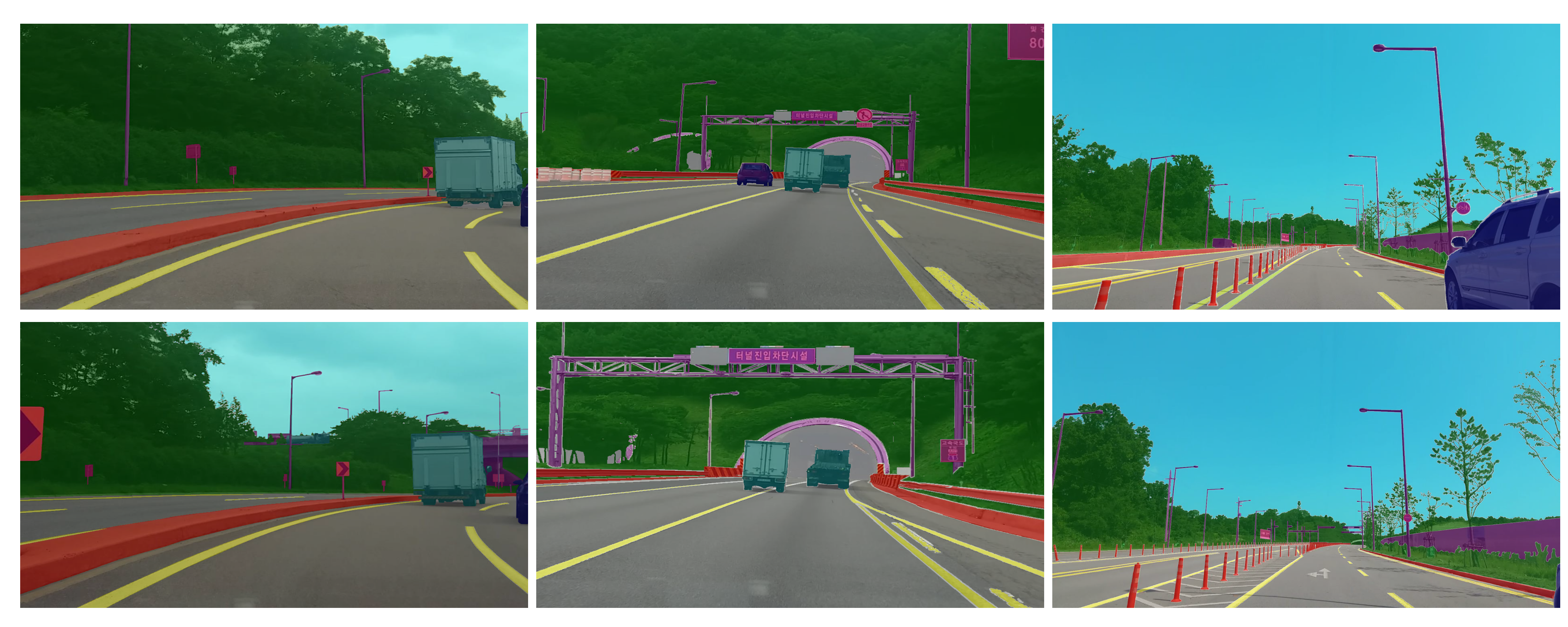}
\caption{Samples from the collected dataset. Each image is overlaid with its annotation. The first row presents the first frames of sequences while the second row presents the last frames}
\label{fig:example}
\vskip -0.4mm
\end{figure}

The underestimation of the runtime constraint is related to the lack of a semantic video segmentation dataset with temporally dense annotation as research progress depends greatly on the existence of datasets~\cite{cityscape}.
In the field of semantic segmentation, there exist well-annotated datasets, such as the Cityscape~\cite{cityscape}, KITTI Vision Benchmark Suite~\cite{kitti}, Daimler Urban Segmentation~\cite{dus}, CamVid~\cite{camvid}, PASCAL VOC~\cite{pascal}, and Microsoft COCO~\cite{coco} datasets. 
These datasets include various scenes, such as indoor, outdoor, office, urban, or driving scenes.
However, the common shortcoming of these datasets is that of the temporal density.
Although the CamVid dataset provides annotated frames at 1Hz, we argue that it is still insufficient for autonomous driving.
Therefore, we introduce the \textit{Highway Driving dataset}, which is spatially and temporally densely annotated.\footnote{The dataset is available at https://sites.google.com/site/highwaydrivingdataset/}
Short video clips with frame rate of 30Hz were captured under a highway-driving scenario.
Every frame of each clip was then densely annotated.
Moreover, each frame is annotated while considering the correlation between the adjacent frames.
They were annotated sequentially, such that the annotation was consistent.
We also propose a baseline algorithm for semantic video segmentation using our dataset.
The major objective of our algorithm is to label the driving scene with a limited time budget.
Therefore, the provided baseline algorithm focuses more on time efficiency.
In other words, the objective is to predict the pixel-level scene labels much faster then existing algorithms with a comparable performance.

The remainder of this paper consists of five additional sections.
In the following section, we introduce the previous research related to our work.
We present the Highway Driving dataset in Section~\ref{section:dataset}.
In Sections~\ref{section:algorithm} and \ref{section:exp}, we introduce our baseline algorithm and its experimental results with the Highway Driving dataset, respectively.
We then conclude this paper in the last section.

\section{Related Works}
In this study, we introduce a new dataset for driving scenarios and provide a baseline algorithm for fast inference in video datasets. 
In this section, we attempt to prove the necessity of the new dataset by exploring works related to the present research.

{\bf Semantic Segmentation Datasets:} While PASCAL VOC~\cite{pascal} and Microsoft COCO~\cite{coco} provide semantic segmentation labels for objects, our paper is focused on driving scenarios. 
Recent works~\cite{cityscape,kitti,dus,camvid,leibe2007dynamic,xie2016semantic} have been focused on building a segmentation dataset for various environments. 
KITTI Vision Benchmark Suite~\cite{kitti} recorded 6 hours of traffic scenarios and provides the 3D and 2D annotations for five categories. 
CamVid~\cite{camvid} offers pixel level annotations of over 700 images at 1Hz in driving scenarios. 
Furthermore, Leuven~\cite{leibe2007dynamic} consists of 3D segmentation labels of 1175 image pairs. 
Huge 3D-2D pair datasets for traffic environments can be found in a paper written by Xie \textit{et al.}~\cite{xie2016semantic}.
%More recently, huge dataset Cityscape~\cite{cityscape} captured in streets from 50 different cities. 
More recently, a large dataset Cityscape~\cite{cityscape} has been presented, which comprises street scenarios from 50 different cities.
This dataset also provides high quality pixel-level annotations for 5000 images. 
%However, this dataset still has disadvantage that has limited to the single image. 
%It cannot be directly used for the video input. 
However, this dataset still has the disadvantage that its application is limited to a single image; it cannot be directly used with a video input.
Therefore, our proposed dataset is essential and is a unique dataset that provides accurate pixel-level annotations for video frames of 30Hz such that it dataset can be used to train deep neural networks (DNNs) for a video input.

{\bf Semantic Segmentation algorithms:} Recent achievements in semantic segmentation have resulted from improvements in DNNs. 
As DNNs provide great performance in image classification task~\cite{resnet,vgg,han2017deep,huang2017densely}, several researchers~\cite{fcn,dilation,deeplab} use pretrained DNNs for segmentation tasks. 
As pretrained DNNs have a small spatial size feature maps in high layers, researchers~\cite{noh2015learning} use the deconvolutional layers after the last layer of the pretrained DNNs such that DNNs can make pixel-level predictions. 
There are research~\cite{dilation,deeplab} that proposed the dilated convolutional layer that can make large spatial size feature maps while using the weights of pretrained DNNs. 
Although they have the drawback that the dilated convolution layer requires a large amount of memory, this approach helps DNNs to have various important information in the high layers.
Furthermore, FCN~\cite{fcn} uses not only the high level features but also the low-level features to achieve fine-level prediction. 
Similarly, PSPnet~\cite{pspnet} uses the pyramid pooling module to predict coarse and fine level prediction.

Despite the absence of a dataset for semantic video segmentation, several algorithms have been proposed in the literature recently~\cite{NextSegmPredICCV17,BudgetAwareDS,DBLP:journals/corr/NilssonS16,7780714}.
Luc \textit{et al.}~\cite{NextSegmPredICCV17} proposed an algorithm for semantic video segmentation using self-supervision.
By predicting the future frames, the network can learn the context of the data without video annotation.
There was also an attempt to utilize the optical flow by Nillson \textit{et al.}~\cite{DBLP:journals/corr/NilssonS16}.
These various approaches have successfully improved the scene labeling performance.
However, these algorithms are rather heavy, so that they require a long time budget.
In contrast, Mahasseni \textit{et al.}~\cite{BudgetAwareDS} proposed an algorithm that emphasizes the time budget.

{\bf Fast inference algorithms:} Several researchers have proposed various methods to allow DNNs to infer faster.
Han \textit{et al.}~\cite{prun} proposed pruning methods that erase less-meaningful parameters of DNNs such that the network can contain a small number of parameters. 
Furthermore, by using weight compression methods~\cite{compress1,compress2}, DNNs can also infer an input signal quickly. 
In particular, the weight binarization method~\cite{binarized1}, which is an extreme case of the weight compression method, can be used to compress the network under the 4\%. 
In addition, the knowledge distillation methods~\cite{fitnet,dark} can also be used to reduce the size of a network by distilling the knowledge from the high-performance network to the small network.

\section{Dataset for Semantic Video Segmentation}
\label{section:dataset}
An autonomous driving system is a real-world application that would greatly require reliability.
Unexpected incidents may occur at any time, and they require immediate, and yet appropriate, responses.
In order to make a good decision, the system should fully understand the situation.
Unfortunately, existing datasets do not contain sufficient information that the system requires.
Some of the dataset provide annotations of independent images~\cite{cityscape,kitti}, and the others provide temporally sparse annotations~\cite{camvid,dus}.
In the following subsection, we introduce the Highway Driving dataset and describe its annotation procedure.

\subsection{Annotations}
The dataset consists of 20 sequences of 60 frames with a 30Hz frame rate.
Therefore, we provide a total of 1200 frames with annotations.
Originally, longer clips were recorded, and we trimmed 2 seconds from each clip.
As we believe that a correlation between adjacent frames is key information in semantic video segmentation, every sequence is carefully annotated while maintaining consistency.
The frames in a single sequence were annotated in chronological order.
Each annotator was asked to annotate adjacent frames, and formerly annotated results for prior frames were provided as their reference.
On average, 2.2 annotators had annotated a single sequence.

The provided annotations are spatially dense as well.
In order to build spatially dense annotations, we had annotated through adversarial procedure.
If an annotator completes the annotation, another annotator identifies defects in it.
This procedure was repeated until they could not find defects in the produced annotation.
Including the time for quality control, annotating a single image requires over 1 h on average.
Using the adversarial annotating procedure, we have obtained spatially dense pixel-level annotations.
Table~\ref{table:stat} presents the spatial~\cite{cityscape} and temporal density of annotations for several driving scene datasets.
Except in the case of the DUS dataset~\cite{dus}, the spatial densities of all the datasets are comparable.
In the case of our dataset, the majority of non-annotated pixels come from the bonnet of the data collecting vehicle.
On the other hand, the temporal density shows greater significance.
Although some other datasets provide video frames, fine annotations are deficient in terms of temporal density.
This shows that our dataset is compactly annotated as compared to other widely used datasets~\cite{cityscape,camvid,dus}.

\subsection{Classes and Evaluation}
\label{section:evaluation}
We defined 10 classes that commonly appear in driving scenarios: road, lane, sky, fence, construction, traffic sign, car, truck, vegetation, and unknown.
The unknown class includes undefined objects, the bonnet of the data-collecting vehicle, and ambiguous edges.
The most relevant classes to autonomous driving were selected from the high-speed driving standpoint.
As the majority of selected classes have an intuitively interpretable definition, we only define some classes here.
The \textit{lane} class is literally the lane on the road.
Other marks printed on the road in order to inform the drivers are excluded.
We define the \textit{fence} as the structures on both side of the road.
The fence class can be considered as a sub-class of the construction class.
However, we have separated this class from the construction class as the fence class is one of the most notice-worthy structures observed during driving.
In Figure~\ref{fig:example}, the fence class is in red.
The \textit{construction} class contains every man-made structure except for the road and fence.
It is indicated in purple in Figure~\ref{fig:example}.
More detailed information regarding the dataset can be found in the supplementary material.

In order to evaluate the labeling performance for each class, we use the intersection over union (IoU)~\cite{pascal}.
A pixel that is annotated as an \textit{unknown} class is not considered as a performance measure.
However, as a performance measure for the whole dataset, the IoU metric is considerably biased to certain classes that cover a large area.
That is undesirable as the classes covering a relatively small area are \textit{not} less important.
Therefore, as a metric for the whole dataset, we use the mean IoU (mIoU)~\cite{cityscape}, which is the IoU averaged over all the classes, so that every class contributes equally to the performance measure.

We split the dataset into training and test sets.
The training set consists of 15 sequences while the test set consists of the remaining five sequences.
Rather than randomly splitting the sequences, we split the training and test sets to have a similar class distribution.
More detailed statistics are presented in the supplementary material.

\begin{table}[]
\small
\centering
\label{table:stat}
\begin{tabular}{lcc}
\hline
               & \multicolumn{1}{l}{spat. density{[}\%{]}} & \multicolumn{1}{l}{temp. density[Hz]} \\ \hline
HighwayDriving & \textbf{97.8} & \textbf{30} \\
CityScapes     &     97.1      &    -        \\
CamVid         &     96.2      &    1        \\
DUS            &     63.0      &    3        \\ \hline
\end{tabular}
\caption{Spatial and temporal density of annotations for the Highway Driving, CityScapes, CamVid, and DUS datasets}
\end{table}

\section{Baseline Algorithm}
\label{section:algorithm}
\begin{figure}[t]
\centering
\includegraphics[width=0.85\textwidth]{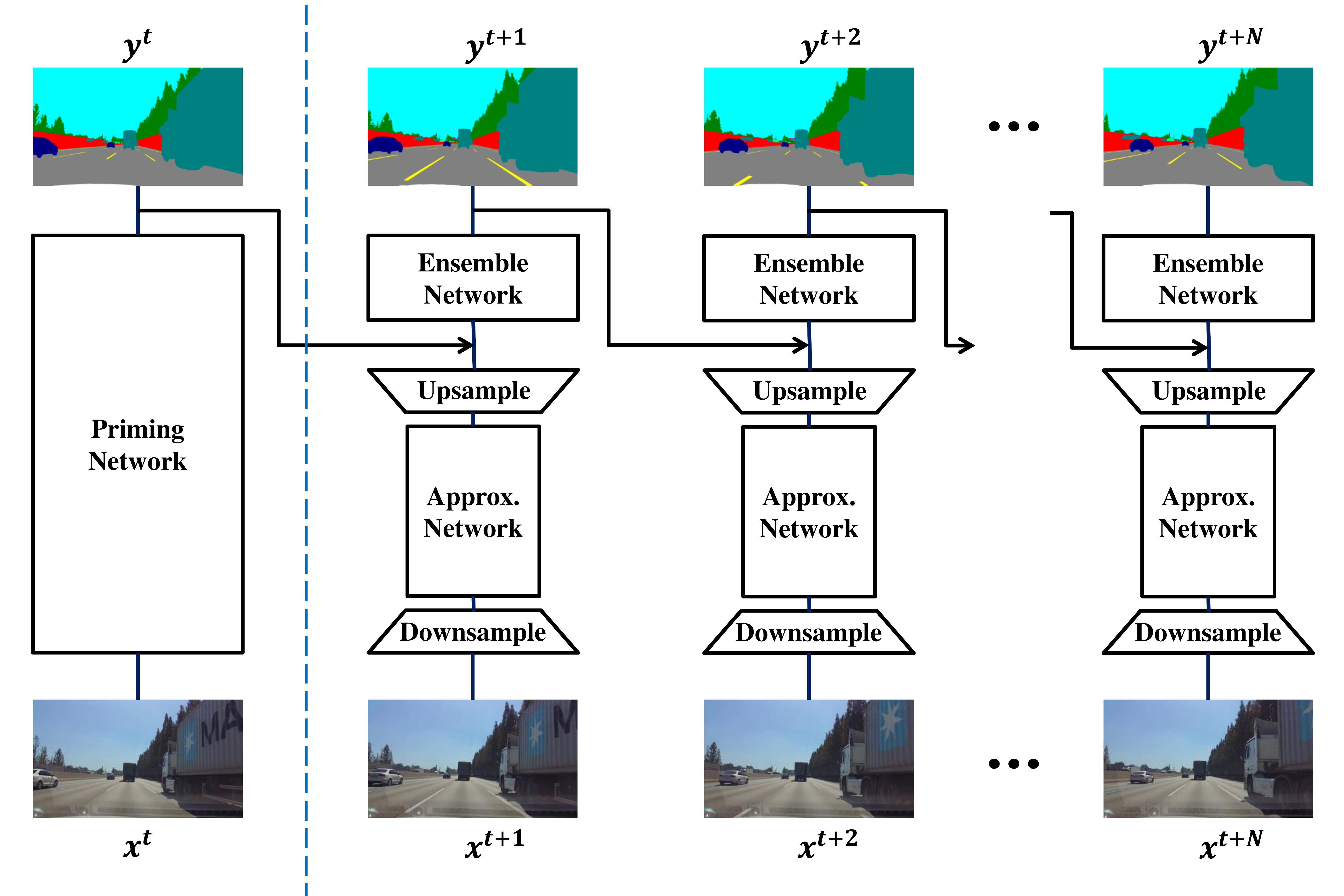}
\caption{Overall architecture of the proposed baseline algorithm. Three different networks recurrently predict the scene labels}
\label{fig:algorithm}
\vskip -4mm
\end{figure}

In this section, we present a baseline algorithm for the Highway Driving dataset, which utilizes a temporal correlation to reduce the prediction time.
As a video is a collection of sequential images, the adjacent frames are highly correlated.
The more we emphasize on the real-time computation, the shorter the time budget we can use, and thus, the correlation between the adjacent frames is the key information for achieving a high performance.
We propose the use of a simple architecture that combines information from two adjacent frames such that the scene labels can be predicted with a limited time budget.
The overall architecture of our algorithm is illustrated in Figure~\ref{fig:algorithm}.
The entire system consists of three networks: the priming network, transition network, and approximating network.
Except for the initial frame, our algorithm recurrently predicts the scene label without the priming network.

{\bf Priming Network} The priming network is a relatively larger and deeper network for the initial frame.
It is simply an image segmentation network that uses any temporal relations between adjacent frames.
As compared to other networks, the priming network can more accurately predict the scene labels while it requires a longer time budget.
The major role of the priming network is to generate and deliver an accurate \textit{prior} knowledge to the transition network for the following frames.
As the priming network initializes the entire system, the performance of the priming network directly affects the performance of the entire algorithm.
In addition, we can control the priming frequency depending on the time budget.

{\bf Approximating Network}
The approximating network is designed to approximately segment the current frame.
We assume that every motion is continuous and smooth.
This may be an incomplete argument under non-driving scenarios; a person can appear to be opening a door, or an object may emerges from behind an obstacle.
However, the majority of the counterexamples in the dice are the urgent situations in driving scenarios, and we believe they require a specialized alert system.
From this perspective, the approximating network should provide a rough scene labeling result, so that the following ensemble network can finalize the fine scene labeling.

In order to shorten the runtime of the approximating network, we downsample the input frame and feed it into the approximating network.
Here, there exists a trade-off between the output performance, which benefits from a high resolution, and the runtime.
From the viewpoint of scene labeling performance, there is no harm in feeding high-resolution images.
The images with a high resolution naturally contain more information and the objects have sharper edges.
However, we still downsample the input frame because we put a greater emphasis on the runtime constraint.
The same trade-off exists for the priming network; the runtime is more weighted for the approximating network while the performance is for the priming network, and thus, we use a wider and deeper architecture for the priming network.

{\bf Ensemble Network}
The ensemble network is a shallow network that transforms the knowledge from the former frame and ensemble the two information from adjacent frames.
This network is required to be extremely thin and shallow as it takes the full-sized scene labeling result as its input.
Despite its small network size, it plays a key role in utilizing the correlation between adjacent frames.
Without the ensemble network, the overall algorithm is no different from frame-by-frame scene labeling with one of the approximating network and priming network.

The feature maps from the current frame, which are obtained from the approximating network, should be upsampled as the frame is formerly downsampled. 
We apply bilinear interpolation to the feature maps before it is fed into the ensemble network.
This implies that there exists an upper limit to the approximating network.
This limitation is measured by control experiments in Section~\ref{subsection:exp}.

As the priming network, we use DRN~\cite{drn} which is known to show great performance on scene labeling~\cite{cityscape}.
In advance of the other networks, the priming network is independently trained with all the training frames.
The approximating network and ensemble network are then jointly trained.
Every parameter in both the priming network and approximating network is initialized from the pretrained model using \textit{ImageNet}~\citep{ILSVRC15}, while the parameters in the ensemble network are initialized randomly.

\section{Experimental Results}
\label{section:exp}
\subsection{Control and Baseline Experiments}
\label{subsection:exp}
As the resolution of the input image directly affects the computation time of the network, the most convenient method for resolving the issue of the runtime constraint is to downsample the input frames.
As the upper bounds for the downsampling approach, we evaluate the performance of the subsampled ground-truth labels in advance to the other existing semantic segmentation algorithms.
In Table~\ref{table:baseline}, the methods named sub-$N$ represent the control experiments: the ground-truth labels subsampled with stride $N$.
The subsampled labels, which have a smaller resolution, are upscaled to the original resolution using nearest neighbor interpolation for the purpose of the evaluation.
The performance of the classes that cover a relatively large area, such as road, sky, and vegetation, are less degraded.
Notable deterioration occurs in the case of several classes that cover a small area.
In particular, the lane class is severely deteriorated.
That is a natural consequence as the lane class not only covers a small portion of images but is also thin, which makes it vulnerable to interpolation degradation.
The deterioration of these classes implies that a complementary method is required for the downsampling method.
However, the robustness of the performances in the case of large classes, such as road, sky, or vegetation, underscores another message that downsampled images are still informative.
This result justifies our baseline approach as our algorithm keeps providing coarsely labeled results by downsampling.

\begin{table}[t]
\small
\centering
\begin{tabular}{l|ccccccccc|c|c}
\hline
   &\rotatebox{90}{Road}& \rotatebox{90}{Lane}& \rotatebox{90}{Sky}& \rotatebox{90}{Fence} & \rotatebox{90}{Construction}       & \rotatebox{90}{Traffic sign} &  \rotatebox{90}{Car}  & \rotatebox{90}{Truck} & \rotatebox{90}{Vegetation} & mIoU & \shortstack{relative \\ runtime } \\ \hline
sub-2       &98.9&87.1&99.5&98.2&96.7&95.7&98.3&98.9&98.7&96.9 & -    \\
sub-4       &98.0&77.3&99.2&96.8&94.1&91.8&96.8&98.2&97.7&94.4 & -    \\
sub-8       &96.4&63.1&98.6&94.0&89.7&85.0&94.0&96.6&96.0&90.4 & -    \\
sub-16      &93.5&45.7&97.6&88.8&83.2&73.5&89.1&93.6&93.1&84.2 & -    \\
sub-32      &89.4&28.7&96.1&80.1&74.7&58.3&80.1&88.1&88.5&76.0 & -    \\ \hline
FCN-32s     &83.5&36.6&96.0&59.9&16.6&31.8&12.3&14.9&82.3&48.2 & -    \\
DRN$\dagger$&91.1&45.8&96.1&69.9&21.2&26.9&53.1&69.2&87.8&62.3 & 1    \\
PSPnet      &92.5&50.4&97.0&69.3&17.3&15.4&66.4&69.5&88.2&62.9 & 2.18 \\ \hline
            &87.2&45.4&93.3&66.5&19.0&22.5&54.3&55.7&86.2&58.9 & 0.78 \\
Ours        &83.8&44.9&90.7&63.3&17.7&19.5&55.4&46.3&84.8&56.3 & 0.58 \\
            &83.2&44.8&90.3&62.6&17.4&18.6&55.5&44.4&84.5&55.7 & 0.36 \\ \hline
\end{tabular}
\caption{Quantitative result of control and preliminary experiments for semantic scene labeling. $\dagger$ represents the priming network of our method}
\label{table:baseline}
\end{table}

\begin{table}[t]
\scriptsize
\centering
\begin{tabular}{l|ccccccccccc|c|c}
\hline
   &\rotatebox{90}{Road}& \rotatebox{90}{Building}& \rotatebox{90}{Sky}& \rotatebox{90}{Tree} & \rotatebox{90}{Side-Walk}       & \rotatebox{90}{Car} &  \rotatebox{90}{Column-Pole}  & \rotatebox{90}{Fence} & \rotatebox{90}{Pedestrian}& \rotatebox{90}{Bicycle} & \rotatebox{90}{Sign} & \rotatebox{90}{Class Avg.} & \shortstack{relative \\ runtime } \\ \hline
\cite{bayessegnet}  &80.4&85.5&90.1&86.4&67.9&93.8&73.8&64.5&50.8&91.7&54.6&76.3& $B_{max}$    \\ 
\cite{BudgetAwareDS}&77.1&81.9&86.2&81.7&65.1&88.7&69.3&61.8&49.1&88.2&52.8&72.9& $0.5 \cdot B_{max}$\\ 
\cite{BudgetAwareDS}&60.3&60.1&64.8&56.7&50.3&60.1&46.8&42.3&33.7&59.4&31.6&51.5& $0.1 \cdot B_{max}$\\ \hline
DRN$\dagger$        &92.5&87.5&94.6&86.4&57.1&97.7&67.9&46.0&26.3&83.5&51.8&71.9& 1    \\ 
Ours                &91.3&87.4&93.3&78.0&47.3&97.4&63.7&31.8&25.2&78.8&44.9&66.8& 0.81    \\
                    &91.2&83.3&93.1&76.9&46.5&97.4&63.4&30.8&25.0&78.3&44.0&66.5& 0.35  \\ \hline
\end{tabular}
\caption{Quantitative result of our algorithm with CamVid dataset. $\dagger$ represents the priming network of our method}
\label{table:camvid}
\end{table}

A straight forward approach for the semantic video segmentation is to segment each frame independently.
Under this scheme, every sequence can be disassembled into independent images and the existing semantic image segmentation algorithms~\cite{drn,fcn,pspnet} are easily applicable.
This approach is computationally expensive but has been intensively researched over the recent years.
Table~\ref{table:baseline} shows the quantitative results.
Each method is evaluated without considering the runtime constraint.
Similar to many other CNN-based approaches, they require plenty of data in order to realize a high performance.
Although we provide 1200 images, our dataset is less rich from the viewpoint of variability as compared to other existing datasets~\cite{cityscape,camvid,dus}, because the 60 frames are severely correlated with each other.
The insufficient variability forces algorithms to focus more on the correlation between the frames.

Table~\ref{table:baseline} presents the performance of our algorithm in terms of both scene labeling and runtime.
The relative runtime shows how fast the algorithm can segment each frame.
As we have used DRN as our priming network, we normalized the runtime of each algorithm with the runtime of DRN.
Table~\ref{table:baseline} shows that our algorithm can achieve a comparable result with a short time budget.
Our algorithm is also capable of controlling its runtime by adjusting the priming frequency.
The priming network is a large but slow network.
Therefore, the more frequently we use the network, the longer the time budget that we require while the better performance we can acquire.

\begin{figure}[t]
\centering
\includegraphics[width=\textwidth]{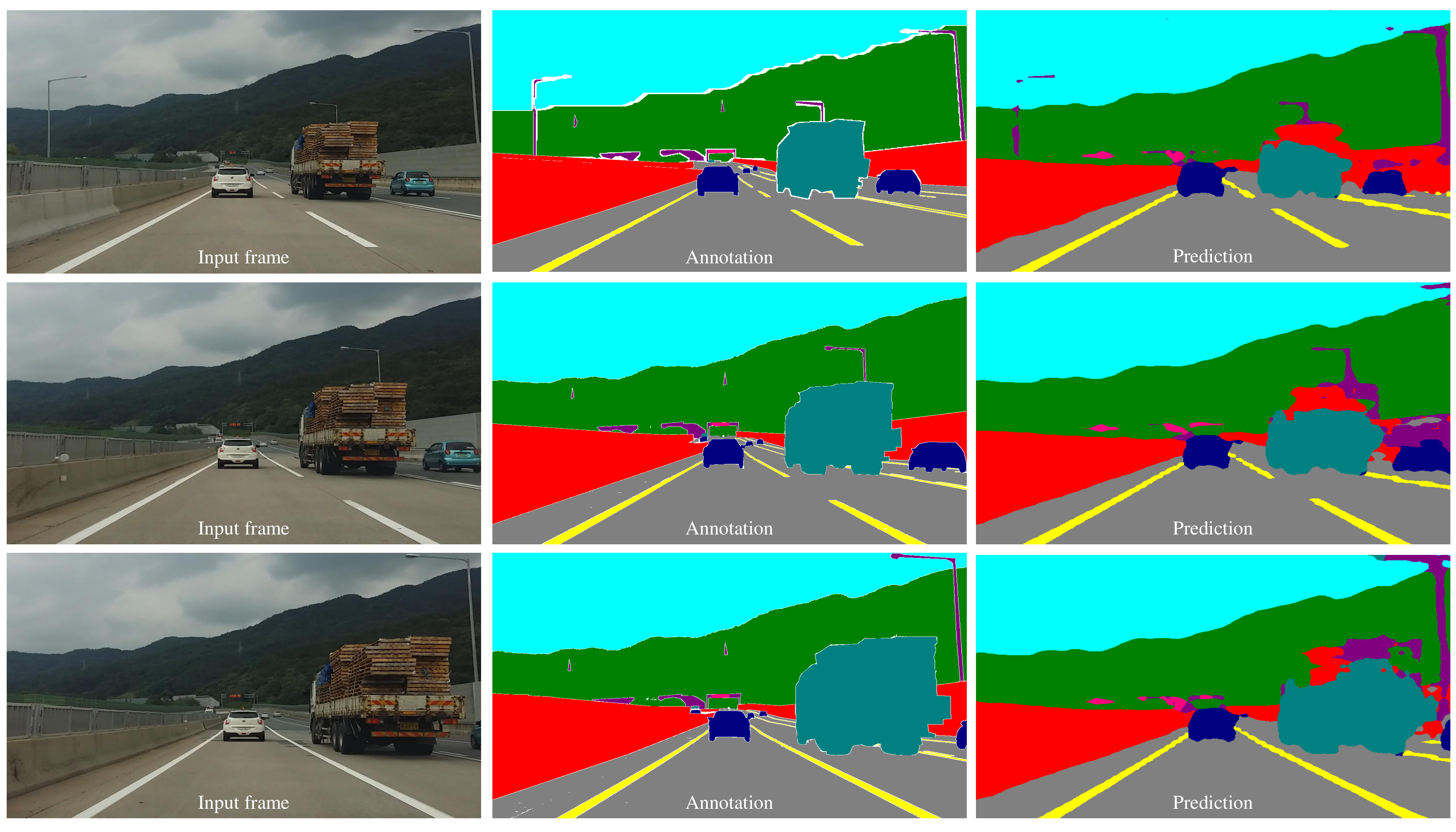}
\caption{Qualitative results of baseline algorithm. From the left to right: image frame, ground truth, and scene labeling result with the baseline algorithm. From top to bottom, the first, middle, and the last frame of the sequence}
\label{fig:result}
\end{figure}

Figure~\ref{fig:result} illustrates qualitative scene labeling result obtained for the Highway Driving dataset.
In the Figure~\ref{fig:result}, only the first frame is labeled with the priming network.
The other frames are labeled with the approximating network and ensemble network.
Since the top row is the first frame of the sequence, the prediction result on the top row is predicted with the priming network, while the other two results are predicted with the approximating network and ensemble network.
As the bottom row represents the 60-th frame, in order to predict the scene label, information from the priming network should come through 59 time steps.
It can still be observed that the results are not severely deteriorated.
This implies that the recurrent framework is operating properly.

Figure~\ref{fig:result} also plainly shows how similar the sequential frames can be.
As the sequence consists of 60 frames, there are another 28 frames between the adjacent rows in Figure~\ref{fig:result}.
Still, the most part of the frames seems identical except in the case of the moving objects.
These objects contain the most important information for autonomous driving as they are the objects with different velocity from the ego-motion.
If the annotations are not temporally dense, it is difficult to localize the moving objects because the ego-motion itself causes large differences between the frames. 
With dense annotations, the ego-motion becomes negligible, so that the network can focus on the moving objects.

\subsection{Experiments on CamVid Dataset}
We evaluated our baseline algorithm using the CamVid dataset \cite{camvid} as a verification. 
The CamVid dataset provides 1Hz annotations over five videos.
We used identical model architectures with the experiment presented in Section~\ref{subsection:exp}.
In addition, we did not apply fine-tuning or additional learning algorithms to improve the performance.
The experimental settings used were based on the prior works \cite{BudgetAwareDS,bayessegnet}.
Table~\ref{table:camvid} shows the performance of our algorithm with the CamVid dataset.
For the CamVid dataset, we have evaluated the performance with the average class accuracy.
As shown in Table~\ref{table:baseline}, we have evaluated the performance with different runtimes by controlling the priming frequency.

Mahasseni \textit{et al.}~\cite{BudgetAwareDS} provided similar experimental results for the CamVid dataset.
Although we cannot directly compare the results obtained due to the difference between the architecture and the time budget used, we can still use their results as a reference for verification.
In the case of both algorithms, natural degradation of the performance is detected as the time budget is reduced.
Table~\ref{table:camvid} shows that our baseline algorithm is not a dataset-specific algorithm.

\section{Conclusion}
\label{section:conclusion}
We introduced the \textit{Highway Driving} dataset-a new benchmark for semantic video segmentation task.
The Highway Driving dataset has significance in temporally dense pixel-level fine annotations.
The provided annotation is denser than other existing datasets in both a spatial and temporal manner.
In addition, we proposed a baseline algorithm for the Highway Driving dataset and verified the algorithm using the CamVid~\cite{camvid} dataset.
The algorithm showed that we could predict the scene labels with a short time budget by using the correlation between the adjacent frames.

Prior to the introduction of the Highway Driving dataset, it has been troublesome to study the semantic video segmentation task owing to the lack of temporally dense annotation.
We expect the temporally dense annotation of the Highway Driving dataset to promote various future research on semantic video segmentation.

\section*{Acknowledgements}
This work was supported by the ICT R\&D program of MSIP/IITP, [2016-0-00563, Research on Adaptive Machine Learning Technology Development for Intelligent Autonomous Digital Companion] and the Industrial Convergence Core Technology Development Program(No. 10063172) funded by MOTIE, Korea.

\bibliography{ref}

\end{document}